\def\year{2022}\relax
\documentclass[letterpaper]{article} 
\usepackage{aaai22}  
\usepackage{times}  
\usepackage{helvet}  
\usepackage{courier}  
\usepackage[hyphens]{url}  
\usepackage{graphicx} 
\usepackage{natbib}  
\usepackage{caption} 
\DeclareCaptionStyle{ruled}{labelfont=normalfont,labelsep=colon,strut=off} 
\usepackage[linesnumbered,ruled,vlined]{algorithm2e}

\usepackage{amsmath}
\usepackage{autobreak}
\usepackage{amsfonts}
\usepackage{diagbox}
\usepackage{multirow}
\usepackage{array}
\usepackage{booktabs}
\usepackage{bm}

\newcommand{\nop}[1]{}

\title{A Generative Car-following Model Conditioned On Driving Styles}
\author{Yifan Zhang\textsuperscript{\rm 1}, Xinhong Chen\textsuperscript{\rm 1}, Jianping Wang\textsuperscript{\rm 1}, Zuduo Zheng\textsuperscript{\rm 2}, Kui Wu\textsuperscript{\rm 3}
}
\affiliations{\textsuperscript{\rm 1}City University of Hong Kong, \textsuperscript{\rm 2}The University of Queensland, \textsuperscript{\rm 3}University of Victoria\\
\{yif.zhang, xinhong.chen\}@my.cityu.edu.hk, jianwang@cityu.edu.hk,  zuduo.zheng@uq.edu.au, wkui@uvic.ca
}

\begin{document}

\maketitle

\begin{abstract}
Car-following (CF) modeling, an essential component in simulating human CF behaviors, has attracted increasing research interest in the past decades. This paper pushes the state of the art by proposing a novel generative hybrid CF model, which achieves high accuracy in characterizing dynamic human CF behaviors and is able to generate realistic human CF behaviors for any given observed or even unobserved driving style. Specifically, the ability of accurately capturing human CF behaviors is ensured by designing and calibrating an Intelligent Driver Model (IDM) with time-varying parameters. The reason behind is that such time-varying parameters can express both the inter-driver heterogeneity, i.e., diverse driving styles of different drivers, and the intra-driver heterogeneity, i.e., changing driving styles of the same driver. The ability of generating realistic human CF behaviors of any given observed driving style is achieved by applying a neural process (NP) based model. The ability of inferring CF behaviors of unobserved driving styles is supported by exploring the relationship between the calibrated time-varying IDM parameters and an intermediate variable of NP. To demonstrate the effectiveness of our proposed models, we conduct extensive experiments and comparisons, including CF model parameter calibration, CF behavior prediction, and trajectory simulation for different driving styles.

\end{abstract}

\section{Introduction}
Car following (CF) is a primary driving task 
that drivers routinely perform~\cite{lc_ZHENG201416,HF2014379}. Thus, realistically modeling human CF behaviors is the foundation of traffic operations and control. It also plays an important role in other traffic related topics, such as traffic safety, vehicle emissions, etc. One common goal of CF behavior modeling is to simulate dynamic human CF behaviors, which are inherently coupled with different driving styles. To this end, an ideal human-like CF model shall be able to: 1) capture human CF strategies accurately, 2) interpret and differentiate different driving styles, and 3) adapt to changing environments. Over the past $60$ years or so, numerous CF models have been proposed to simulate dynamic human CF behaviors. Nevertheless, none of them meets all the aforementioned three requirements. 

Existing CF models can be roughly grouped into three categories: mathematical CF models, data-driven CF models, and hybrid CF models. A mathematical CF model is defined as a pre-designed mathematical form with a set of parameters~\cite{IDM,OVM,GIPPS1981105}. Although this type of CF models has high interpretability through the parameters that have concrete physical meanings~\cite{coeff_corr}, the limited number of parameters in the pre-defined functions or rules makes it hard to characterize human CF strategies accurately in different interactive traffic environments.

The data-driven CF models apply advanced machine learning methods, such as recurrent neural networks~\cite{lstm2020102785,lstm2018car}, reinforcement learning~\cite{rl2018348}, and imitation learning~\cite{il20185034}. Such models can adapt to changing environments and have a high prediction accuracy in learning human CF behaviors. Nevertheless, this type of CF models generally suffers from poor interpretability.

The hybrid CF models~\cite{pid8424196,pid2021103240} leverage the interpretability of mathematical CF models and the high prediction accuracy and adaptability of data-driven CF models. In particular, a mathematical CF model is first calibrated to generate multiple unobserved data samples for training a data-driven CF model. However, the mathematical CF models in the existing hybrid CF models neglect the fact that each driver may change their CF strategies when encountering different environments over time. Utilizing inaccurate CF strategies would result in the poor quality of the generated training samples and decreases the prediction performance of the data-driven CF model.


This paper advances the literature by proposing a novel generative hybrid CF model, which meets all requirements of an ideal human-like CF model. It considers the dynamic nature of human drivers' driving styles to improve the accuracy of the captured human CF strategies. Besides, it is also capable of generating realistic human CF behaviors for any given driving style. The proposed hybrid CF model is constructed based on the integration of an Intelligent Driver Model (IDM) with time-varying parameters and a neural process based (NP-based) CF model. The main contributions of the paper are summarized as follows.

\begin{itemize}
    
    \item To accurately capture dynamic human CF strategies, the inter-driver heterogeneity and the intra-driver heterogeneity are modeled by designing and calibrating an IDM with time-varying parameters. Modeling each parameter as a stochastic process and calibrating them in a time-varying manner enable us to capture the stochasticity and the randomness of human CF behaviors. 
    
    
    
    
    \item To generate realistic human CF behaviors for any observed driving style, an NP-based CF model is proposed and applied accordingly. The stochasticity of the NP makes it capable of taking the randomness caused by uncertain human perceptions and psychological fluctuations into consideration. Thus, realistic and dynamic human CF behaviors can be generated.  
    
    
    \item To infer CF behaviors of unobserved driving styles, an intermediate variable of NP is interpreted as the driving style vector and associated with an explainable and continuous aggressiveness index derived from the calibrated time-varying IDM parameters. By taking such an intermediate variable with concrete physical meanings as input, the well-trained NP-based CF model can infer CF behaviors of any driving style as long as its corresponding aggressiveness index is given.
    
    \item Extensive experiments have been conducted to validate the effectiveness of the proposed models, and the results show that our models achieve $74\%$ increase in the calibration accuracy and $2\%$ accuracy improvement with a much smaller variance in the CF behavior prediction compared to the current state of the art. The above results and a detailed analysis towards simulated trajectories demonstrate that our NP-based CF model is capable of generating various types of realistic traffic flow consisting of both observed and unobserved driving styles.
\end{itemize}


\section{Calibrate IDM With Time-varying Parameters}
Most mathematical CF models describe CF behaviors using a pre-designed mathematical form with a set of predefined parameters, while these parameters should not be fixed in order to accommodate the driver heterogeneity~\cite{IDMM,desired_thw0094351,desired_thw2018194,ltst2020102698}. Thus, we propose to calibrate these parameters in a time-varying manner for each driver. To describe our proposed calibration method, we select IDM as a representative of mathematical CF models, since IDM has been proved to be capable of characterizing human drivers' CF behaviors reasonably~\cite{IDM_best_ZHU2018425}. 

\textbf{General Calibration.}
The acceleration function of a typical CF model is formulated as $a_n = f_\theta(X)$, where $\theta$ is a vector of CF model parameters, $X$ is a given traffic condition, and $a_n$ is the acceleration of the following vehicle under $X$. To calibrate the parameters of a CF model using the trajectory data, a general calibration problem~\cite{PUNZO2021103165,calibration_OSORIO2019156} is formulated as:
\begin{equation}
\small
\begin{aligned}
    \min \mathcal{J} & (\bm{MoP}^{obs}, \bm{MoP}^{sim}) \\
    \mathrm{s.t. } & \  \bm{MoP}^{sim} = F(f_\theta(\mathbf{X})) \\
    & \ LB_\theta \leq \theta \leq UB_\theta
\end{aligned}
\end{equation}
where $MoP$ is the \textit{measure of performance}, e.g., acceleration, spacing, and speed; $MoP^{obs}$ and $MoP^{sim}$ are the observed and the simulated $MoP$, respectively; $F$ is used to compute $MoP$ given the simulated acceleration; $\mathcal{J}(\cdot)$ is a \textit{goodness-of-fit function (GoF)}, e.g., root mean squared error (RMSE) and mean absolute error (MAE); $\mathbf{X}$ is\nop{the} a set of observed traffic conditions extracted from the trajectory data; $LB_\theta$ and $UB_\theta$ are the lower and the upper bounds of the parameters $\theta$, respectively. \nop{The objective of CF model calibration is to minimize $\mathcal{J}(MoP^{obs}, MoP^{sim})$.}The notations used to describe the calibration process are given in Table~\ref{tab:notations}. We adopt the following ballistic integration scheme to update $v_n$\nop{vehicle speed} and $x_n$\nop{position using the simulated acceleration}.
\begin{equation} \label{eq:veh_update}
\small
\begin{aligned}
    v_n(t+\Delta t) & = v_n(t) + a_n(t)\cdot \Delta t \\
    x_n(t+\Delta t) & = x_n(t) + \frac{v_n(t+\Delta t) + v_n(t)}{2} \cdot \Delta t 
\end{aligned}
\end{equation}

\begin{table}[t]
    \small
    \centering
    \begin{tabular}{lm{6.2cm}}
    \toprule
    \textbf{Notation} & \textbf{Description} \\
    \midrule
    $a_n(t)$ & Acceleration of the vehicle $n$ at time $t$ \\
    \specialrule{0em}{1pt}{1pt}
    $s_n(t)$ & Spacing between vehicle $n$ and its leading vehicle at time $t$ \\
    \specialrule{0em}{1pt}{1pt}
    $x_n(t)$ & Position of the vehicle $n$ at time $t$ \\
    \specialrule{0em}{1pt}{1pt}
    $v_n(t)$ & Speed of the vehicle $n$ at time $t$ \\
    \specialrule{0em}{1pt}{1pt}
    $\Delta v_n(t)$ & Relative speed of the vehicle $n$ at time $t$, i.e., $v_{n}(t) - v_{n-1}(t)$ \\
    \specialrule{0em}{1pt}{1pt}
    $\theta_t$ & CF parameters at time $t$\\
    \specialrule{0em}{1pt}{1pt}
    $n$ & Following vehicle \\
    \specialrule{0em}{1pt}{1pt}
    $n-1$ & Leading vehicle \\
    \bottomrule
    \end{tabular}
    \caption{Notations used in the calibration process.}
    \label{tab:notations}
\end{table}

\textbf{IDM with Time-varying Parameters.}
The mathematical formulation of the original IDM acceleration function is presented as follows.
\begin{equation} \label{eq:IDM}
\fontsize{8pt}{9.5pt} \selectfont
    \begin{aligned}
        & a_n(t) = a_{max, n} \cdot \left[ 1 - \left( \frac{v_n(t)}{v_{0, n}} \right) ^\delta - \left( \frac{s^*(v_n(t), \Delta v_n(t))}{s_n(t)} \right) ^2 \right] \\ 
        & s^*(v_n(t), \Delta v_n(t)) = s_{0,n} \\
        & \qquad \qquad \qquad \quad + \max \left[ 0, \left( v_n(t) \cdot T_n + \frac{v_n(t) \cdot  \Delta v_n(t)}{2\sqrt{a_{max, n} \cdot b_n}} \right) \right]
    \end{aligned}
\end{equation}
The parameters to be estimated include $v_{0,n}$, $T_n$, $s_{0,n}$, $a_{max,n}$, and $b_{n}$, which are desired speed, desired time headway, standstill distance gap, maximum acceleration, and desired deceleration of the vehicle, respectively. The exponent $\delta$ is commonly fixed as $4$~\cite{IDM}. $s^*(v_n(t),$ $\Delta v_n(t))$ is the desired spacing and can be computed according to Equation~(\ref{eq:IDM}). 

To tackle the driver heterogeneity, we define an IDM with time-varying parameters (as shown in Equation~(\ref{eq:tvidm})) where the vector of dynamic parameters at time $t$ is denoted as $\theta_n(t)=(v_{0,n}(t),$ $T_n(t),$ $s_{0,n}(t),$ $a_{max,n}(t),$ $b_n(t))$ and abbreviated as $\theta_t$. Specifically, the inter-driver heterogeneity is reflected by calibrating different CF models for different drivers, while the intra-driver heterogeneity is captured by such time-varying parameters of a driver's CF model. 
\begin{equation}
\label{eq:tvidm}
\fontsize{8pt}{9.5pt} \selectfont
    \begin{split}
        & a_n(t) = a_{max, n}(t) \cdot \left[ 1 - \left( \frac{v_n(t)}{v_{0, n}(t)} \right) ^\delta - \left( \frac{s^*(v_n(t), \Delta v_n(t))}{s_n(t)} \right) ^2 \right] \\ 
        & s^*(v_n(t), \Delta v_n(t)) = s_{0,n}(t) \\ 
        & \qquad \qquad \qquad + \max \left[ 0, \left( v_n(t) \cdot T_n(t) + \frac{v_n(t) \cdot  \Delta v_n(t)}{2\sqrt{a_{max, n}(t) \cdot b_n(t)}} \right) \right]
    \end{split}
\end{equation}

\textbf{Calibrate Time-varying IDM Parameter.}
We design a new calibration process to calibrate the time-varying parameters defined in Equation~(\ref{eq:tvidm}) for an individual driver.
\begin{equation}
\small
\begin{aligned}
    & \Theta^* = \mathop{\arg\min}\limits_{\Theta} \sum_{t \in \mathcal{T}} \mathcal{J} (MoP^{obs}(t), MoP^{sim}(t)) \\
    \mathrm{s.t.}     & \ MoP^{sim}(t) = F(f_{\theta_t}(X_t)) \\ 
    & \ LB_{\theta} \leq \theta_t \leq UB_{\theta}, t\in \mathcal{T}
\end{aligned}
\end{equation}
where $X_t$ is the traffic condition at time $t$, and $MoP^{obs}(t)$ and $MoP^{sim}(t)$ are the observed and the simulated $MoP$ at time t, respectively.
Considering the uncertain human perceptions and the inherent randomness in driving behaviors~\cite{sp_NGODUY2019599,sp_LEE2019360}, the time-varying parameters are formulated as stochastic processes $\{\pi(\theta_t), t\in \mathcal{T}\}$. Then, the calibration process aims to estimate the posterior probability distribution of $\{\pi(\theta_t|D), t\in \mathcal{T}\}$ conditioned on a set of observed trajectory data $D$ of an individual driver.

To determine the form of the prior distribution, we apply the multi-model fitting algorithm~\cite{UFLdelong2012fast} on each driver to calibrate multiple sub-models with different sets of parameters to explore the features of the time-varying parameters. From the calibration results, the mean values of the sub-models' parameters calibrated by different subsets of data approximately equal the model parameters calibrated \nop{fitted} using all data. Such an observation implies that a driver has a consistent long-term driving style and varying short-term driving styles that may fluctuate around the long-term one. The long-term driving style is represented by the CF parameters of the model calibrated using all data, and the short-term driving styles are represented by the CF parameters of the sub-models. Moreover, since the driving style of the previous time step has an impact on the current time step, the driving styles of two adjacent time steps are assumed to be similar. For the first time step, the optimal CF parameters should be found near the CF parameters of the long-term driving style. Then, for the rest of the time steps, the optimal CF parameters are supposed to be found near the CF parameters of the previous time step. 

Following the above observations and assumptions, the prior distribution of the first time step is initialized as a Gaussian distribution with the long-term driving style CF parameters as the mean and a predefined hyper-parameter as the variance. For the rest of the time steps, the prior distribution is the posterior distribution of the previous time step. Thus, the time-varying parameters of all time steps are optimized sequentially. To obtain the posterior distribution at each time step, we sample massive sets of parameters from its prior distribution, and only the samples that satisfy the predefined acceptance rules are kept. Once there are sufficient accepted samples, a posterior distribution is fitted using these samples. By analogy, each CF parameter at each time step is modeled as a distribution, and thus the CF parameters of a driver are modeled as stochastic processes.

The detailed calibration process for an individual driver is presented in Algorithm~\ref{al:calibration}, where the interior point (IP) algorithm~\cite{ip2016calibrating} is applied in line 1 for conventional model calibration \nop{in line 1 }using all data of this driver. \nop{parameter optimization.}


\begin{algorithm}[t]
\SetKwInOut{Input}{Input}\SetKwInOut{Output}{Output}
\fontsize{9pt}{8pt} \selectfont
\caption{Time-varying Parameter Calibration}
\label{al:calibration}
\Input{Observed data points $\mathbf{X}^{obs} = \{X^{obs}_t, t \in \mathcal{T}\}$, number of samples $N$, accept thresholds $(\varepsilon, n_{min}, p\%)$, prior distribution covariance diagonal matrix $\Sigma^2$, maximum iterations $M$} 
\Output{Posterior distributions of the time-varying parameters $\{\pi(\theta_t), t \in \mathcal{T}\}$}
\nop{Initialize }$\theta_{prior}^{fix} = \mathop{\arg\min}\limits_{\theta}  \mathcal{J}(\bm{MoP}^{obs}, \bm{MoP}^{sim}),$ $\pi_{prior} = N(\theta_{prior}^{fix}, \Sigma^2)$\;
\For{$t \in \mathcal{T}$}
{
$\Theta^*_{post} = \emptyset$\;
\For{$i = 1:M$}{
Sample $\Theta_{post}=\{\theta_j, j=1,...,N\}$ from $\pi_{prior}$\;
$\Theta^*_{post}=\{\theta_j, \mathcal{J} \left(MoP^{obs}(t), MoP^{sim}(t)\right) < \varepsilon\} \cup \Theta^*_{post}$\;
\If{$\Theta^*_{post}$ is $\emptyset$}
{
\fontsize{8.5pt}{8pt} \selectfont
$\theta_{prior} = \mathop{\arg\min}\limits_{\theta \in \Theta_{post}} \mathcal{J}(MoP^{obs}(t), MoP^{sim}(t))$\;
\fontsize{9.0pt}{8pt} \selectfont
$\pi_{prior} = N(\theta_{prior}, \Sigma^2)$\;
}
\ElseIf{$|\Theta^*_{post}| > n_{min}$}
{
fit the posterior distribution $\pi(\cdot|\Theta^*_{post})$\;
$\pi_{prior} = \pi(\cdot|\Theta^*_{post})$\;
$\Theta^*_{post} = \emptyset$\;
}
\If{$\frac{|\Theta^*_{post}|}{N} > p\%$}
{
$\pi(\theta_t) = \pi_{prior}$\;
\textbf{break}\;
}
}
\If{$\Theta^*_{post}$ is $\emptyset$}
{
$\pi(\theta_t) = N(\theta_{prior}^{fix}, \Sigma^2)$\;
}
\Else
{
$\pi(\theta_t) = \pi_{prior}$\;
}
} 
\end{algorithm}

\section{NP-based Generative CF Model Conditioned On Driving Styles}
Although we can describe human drivers' CF behaviors by calibrating the CF model with time-varying parameters, the generation of such CF behaviors is still challenging since many factors can impact the time-varying CF parameters. Some of these factors are observable, e.g., the cut-in of adjacent vehicles, while others are not, e.g., the psychological fluctuations of drivers. Thus, it is difficult to predict the time-varying parameters.\nop{Thus, it is difficult to model the relations between these factors and the time-varying CF parameters explicitly. As a result, the time-varying CF parameters are hard to predict.} To tackle this issue, we leverage neural networks \nop{based methods }to achieve a\nop{ human-like} CF model that can generate human-like CF behaviors of any \nop{fulfill a }predefined driving style. To meet the stochasticity of human CF behaviors, we propose to approximate the acceleration function of the CF model using neural processes (NPs)~\cite{garnelo2018neural} since NP has remarkable capabilities of function approximation by benefiting from both neural networks and stochastic processes.

\subsection{Basis of Neural Processes}
The objective of NPs is to approximate a function with a neural network that represents a distribution over functions rather than a single fixed function. To this end, the function to be approximated $\mathcal{F}: \mathcal{X} \rightarrow \mathcal{Y}$ is modeled as a stochastic process. Given a finite sequence $x_{1:n}=(x_1, ..., x_n)$, the marginal joint distribution over the function values is $Y_{1:n}:=(\mathcal{F}(x_1), ..., \mathcal{F}(x_n))$. Denoting $z$ as a high-dimensional latent random variable that parameterises $\mathcal{F}$ and introduces the stochasticity into $\mathcal{F}$, we can rewrite $\mathcal{F}(x) = g(x, z)$. Due to the existence of noise and inaccuracy of observations, we let $Y_i \sim N(\mathcal{F}(x_i), \sigma^2)$. Thus, an NP is formulated as a generative model that follows:
\begin{equation}
\small
    p(z, y_{1:n}|x_{1:n}) = p(z)\prod_{i=1}^n N(y_i|g(x_i, z), \sigma^2)
\end{equation}
where $p$ represents the abstract probability distribution.

Following general ideas of generative models, an NP is composed of three parts, namely, encoder, aggregator, and decoder. Given a number of observed data points $(x, y)_i$, we split them into two parts: context points $C=\{(x, y)_i\}_{i=1}^{n_c}$ and target points $T=\{(x, y)_i\}_{i=1}^{n_c + n_t}$, where the target points include the context points. For training, both $C$ and $T$ are used. For testing, $C$ and $x_T=\{x_i\}_{i=1}^{n_c + n_t}$ are provided to predict the target values $\hat{y}_T=f(x_T)$, where $f$ is instantiated from stochastic process $\mathcal{F}$. 
\begin{itemize}
    \item The \textbf{encoder} $h$ is responsible for mapping each context point $(x, y)_i \in C$ to its latent representation $r_i=h((x, y)_i)$.
    \item The \textbf{aggregator} $a$ summarises the output of the encoder. We apply the mean function $r=a(r_i)=\frac{1}{n}\sum_{i=1}^n r_i$ as suggested in~\cite{garnelo2018neural}. 
    \item The \textbf{conditional decoder} $g$ accepts $x_T$ and a sampled $z \sim N(\mu(r), I\sigma(r))$ as input. It outputs the predicted mean and variance of $y_T$ which are denoted as $\mu(\hat{y}_T)$ and $\sigma(\hat{y}_T)$, respectively.
\end{itemize}

\subsection{Model Design}
\begin{figure}[t]
    \centering
    \includegraphics[width=\linewidth]{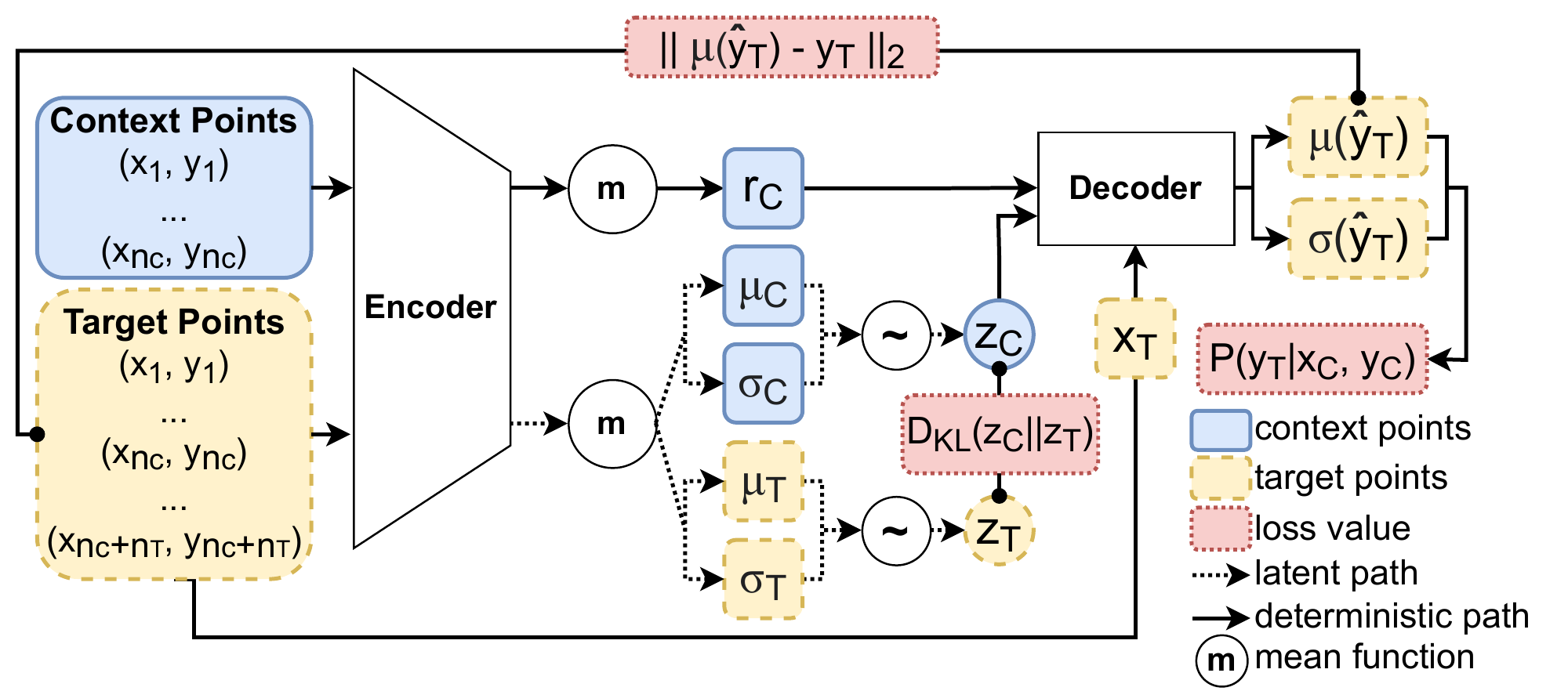}
    \caption{The training process of the NP-based CF model.}
    \label{fig:NP_training}
\end{figure}
\begin{figure}[t]
    \centering
    \includegraphics[width=\linewidth]{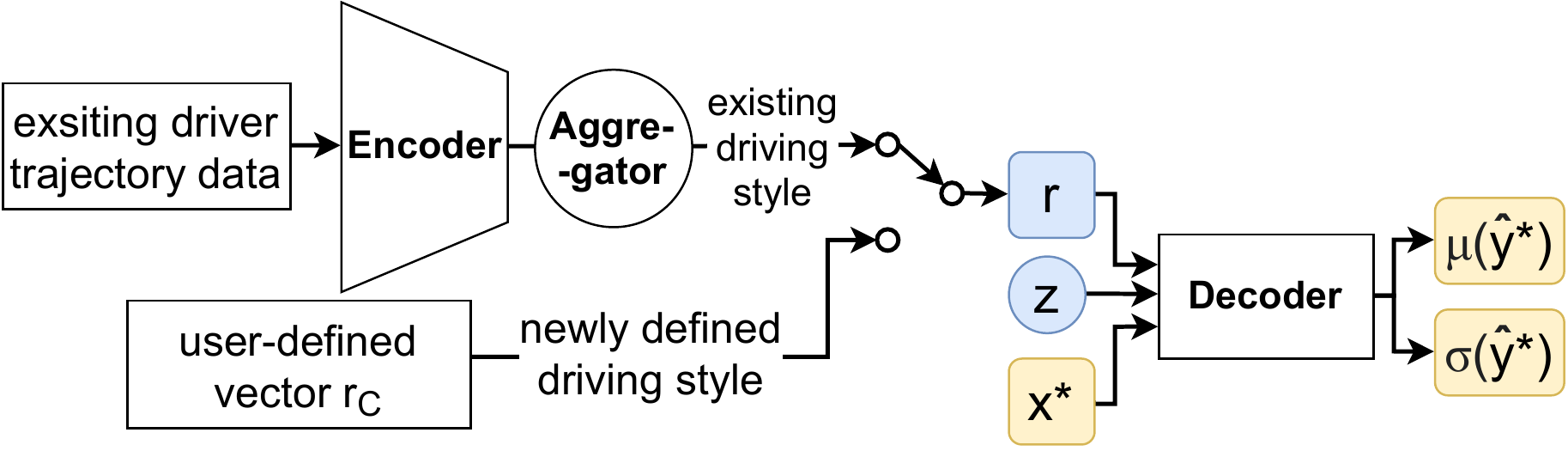}
    \caption{NP-based CF model.}
    \label{fig:NP_test}
\end{figure}
The above NP takes traffic condition $x$ as input and predicts human driver's acceleration $y$ under $x$.
As shown in Figure~\ref{fig:NP_training}, $r_C$ can capture the information from the context points and hence can represent the human driver's driving style.
Next, we elaborate on applying the NP to generate human-like CF behaviors for any given driving style.

\nop{As mentioned above, it mainly includes an encoder, an aggregator, and a conditional decoder.}
\textbf{Model architecture.}
The architecture of the model is illustrated in Figure~\ref{fig:NP_training}. We use a $6$-dimension vector to represent traffic condition $x$ which includes $\Delta v_n(t)$, $v_n(t)$, $s_n(t)$, $v_{n-1}(t)$, $a^{lat}_{n-1}(t)$, and $a^{lon}_{n-1}(t)$. $a^{lat}_{n-1}(t)$ and $a^{lon}_{n-1}(t)$ are the lateral and the longitudinal accelerations of the leading vehicle, respectively. The output $y$ is the following vehicle's acceleration $a_n(t)$. As suggested in~\cite{anp_kim2019attentive}, there are a latent path and a deterministic path in the encoder, denoted as $h_{lat}(\cdot)$ and $h_{det}(\cdot)$, respectively. 
\nop{The output of the latent path brings stochasticity to NP, while the output of the deterministic path aggregates the information of the context points.} 
Specifically, the latent path outputs the mean $\mu_\mathcal{I}= h^\mu_{lat}(\mathcal{I})$ and the variance $\sigma_\mathcal{I} = h^{\sigma}_{lat}(\mathcal{I})$ of latent variable $z$ conditioned on input $\mathcal{I}$, where $\mathcal{I}$ could be a batch of context points or target points. The deterministic path outputs $r \in \mathbb{R}^5$ since five IDM parameters have been proved to be able to characterize typical CF behaviors~\cite{IDM}.

\textbf{Loss function.} As shown in Equation~\ref{eq:loss}, there are three parts in the loss function: (1) the log probabilities of target values $y_T$ over the predicted distribution $N(\mu(\hat{y}_T), \sigma(\hat{y}_T))$; (2) the KL divergence between $h_{lat}(C)$ and $h_{lat}(T)$; and (3) the reconstruction error between $y_T$ and $\mu(\hat{y}_T)$ which is one of the outputs from the decoder\nop{$\mu(\hat{y}_T)$}. The first item needs to be maximized, while the latter two need to be minimized. \nop{Thus, the loss function used to optimize {the?} NP is represented as:}
\begin{equation} \label{eq:loss}
\small
    -\log p(y_T|x_T, C) + D_{KL}(z_C || z_T) + ||\mu(\hat{y}_T) - y_T||_2
\end{equation}
where $z_C$ and $z_T$ are two Gaussian distributions with mean $h^\mu_{lat}(C)$, $h^\mu_{lat}(T)$ and variance $h^{\sigma}_{lat}(C)$, $h^{\sigma}_{lat}(T)$, respectively.

\textbf{Training process.}
The data points from one driver are considered as the sample points from the same stochastic process. For each driver, we split the points into context points and target points. During training, the context points are selected randomly, and the extra target points are selected from the remaining points without replacement. Loss function~(\ref{eq:loss}) is used to optimize all the neural networks\nop{, i.e., the encoder and the decoder}.

\textbf{Testing process.}
After training the NP, \nop{the output of the encoder's deterministic path,} $r$ is found to be similar for the same driver even if the number of context points is different. This implies that $r$ can well capture the driving style of a driver. Therefore, it is reasonable to apply the well-trained decoder as a CF model which accepts traffic condition $x^*$, driving style $r$, and a sampled random variable $z$ as inputs to output the distribution of the corresponding acceleration as shown in Figure~\ref{fig:NP_test}. \nop{The driving style vector can be either encoded from the existing driving trajectory or defined directly by users.}

\section{Explainable Driving Style Representation}
\label{sec:bridge}
So far, we have explainable but unpredictable time-varying IDM parameters, and an unexplainable but generative NP-based CF model. Building a bridge between them can fill their gaps and benefit from both of them. As discussed above, the intermediate variable $r$ can be used to represent the driving style, but the value of $r$ is unexplainable, which makes it difficult to define a CF model with an explainable driving style. Thus, we create a mapping function to associate $r$ with the time-varying IDM parameters that express different aggressiveness levels of drivers.

\begin{table}
\fontsize{8pt}{9pt} \selectfont
    \centering
    \begin{tabular}{p{3.0cm}p{0.5cm}p{0.5cm}p{0.5cm}p{0.8cm}p{0.5cm}}
    \toprule
    \textbf{Parameter} & $v_0$ & $T$ & $s_0$ & $a_{max}$ & $b$ \\
    \midrule
    \textbf{Statistical value} & \textbf{+} & \textbf{--} & \textbf{--} & \textbf{+} & \textbf{+} \\
    \hline
    \textbf{Increasing amplitude} & \textbf{+} & \textbf{--} & \textbf{--} & \textbf{+} & \textbf{+} \\
    \hline
    \textbf{Decreasing amplitude} & \textbf{--} & \textbf{+} & \textbf{+} & \textbf{--} & \textbf{--} \\
    \bottomrule
    \end{tabular}
    \caption{The impact of the parameters on the aggressiveness level. ``+" and ``--" means the potential increase, and decrease of the aggressiveness, respectively.}
    \label{tab:agg_idx}
\end{table}

\textbf{Compute Aggressiveness Index.}
We leverage the calibrated time-varying IDM parameters to compute the aggressiveness index for a driver. Since the parameters may change over time, the aggressiveness level of a driver can be reflected by statistical characteristics and \nop{the }changing amplitudes of the parameters. For example, a higher average value of desired speed means a higher level of aggressiveness. The increased desired speed can also suggest that the driver becomes more aggressive. The mean of $\pi(\theta_t)$ is considered as the parameters at time $t$ for further computation.\nop{Thus, the aggressiveness level is considered from two aspects: statistical value and changing amplitude.}

We use the mean $M(\theta)$, 1/4 percentile $Q1(\theta)$, and 3/4 percentile $Q3(\theta)$ of the time-varying parameters to represent the impacts of statistical value. As for capturing the impacts of changing amplitude, we derive the temporal differential sequence of each parameter, denoted as $\{\theta'(t),$ $t\in \mathcal{T} \}$. Since an increase and a decrease of a parameter indicate opposite impacts on the aggressiveness level, we separate the differential sequence of each parameter as two sub-sequences $\{\theta'^+(t),$ $t|\theta'(t)>0\}$ and $\{\theta'^-(t),$ $t|\theta'(t)<0\}$. After being scaled to $[0, 1]$, the mean $M(\cdot)$ and the standard deviation $S(\cdot)$ of these two sub-sequences are used to represent their impacts on the aggressiveness level. Table~\ref{tab:agg_idx} shows the different impacts of different parameters on the aggressiveness level, where ``+" indicates a positive impact and ``--" indicates a negative impact. Thus, the aggressiveness, denoted as $H \in \mathbb{R}^1$, is computed using Equation~(\ref{eq:agg_idx}).
\begin{equation} 
\label{eq:agg_idx}
\fontsize{8.5pt}{9pt} \selectfont
\begin{aligned}
    H = & \sum_{i=1}^5 \left[ R_i \cdot \left( M(\theta)_i + Q1(\theta)_i + Q3(\theta)_i + M({\theta'}^+)_i + S({\theta'}^+)_i\right. \right. \\ 
    & \left. \left. \qquad \qquad - M({\theta'}^-)_i - S({\theta'}^-)_i \right) \right] , \\
    \mathrm{with}\ &R = [+1, -1, -1, +1, -1], \theta = [v_0, T, s_0, a_{max}, b]
\end{aligned}
\end{equation}

\textbf{Map Aggressiveness Index to $r$.}
With the computed aggressiveness index, the next step is to create a mapping function between the aggressiveness index $H$ and the intermediate variable $r$ of the NP. 
By doing so, once provided with an aggressiveness index $H$ and traffic condition $x$, the decoder of the NP can predict the corresponding acceleration. For each driver, we use all of the data points as the context points to obtain the driving style vector $r$ of this driver. Thus, we have $\{r_i, i\in[1, ..., n]\}$ and $\{H_i, i\in[1, ..., n]\}$, where $n$ is the number of drivers. To simplify the relationship between each pair of $r_i$ and $H_i$, we apply principal component analysis (PCA) on $r$ to reduce its dimension. The reduced $r$ is denoted as $\tilde{r}=\{\tilde{r}_i, i\in [1,...,n]\}$, where $\tilde{r}_i \in \mathbb{R}^1$. Thus, we can fit a function $\tilde{r}_i=f_{map}(H_i)$ to calculate $\tilde{r}_i$ given aggressiveness index $H_i$. To reconstruct $r_i$, inverse PCA is required using the principal components $W$ and the vector of empirical means $u$ as illustrated in Equation~(\ref{eq:map_agg_2_r}). 
\begin{equation} \label{eq:map_agg_2_r}
\small
    \hat{r}_i = f_{map}(H_i) W + u
\end{equation}
\nop{The feasibility of the proposed mapping function~(\ref{eq:map_agg_2_r}) is demonstrated using the real dataset in the following section.}

\section{Experiments}
We conducted numerous experiments to validate the feasibility of the mapping function~(\ref{eq:map_agg_2_r}) and evaluate our calibration algorithm and CF model from the perspectives of calibration performance, prediction performance, and simulated trajectory analysis. The experiment results indicate that the proposed IDM with time-varying parameters achieves the best performance in calibration, and the proposed NP-based CF model can give the most accurate and stable predictions. Also, the simulated trajectories of the observed driving styles closely match the ground truth, while the simulated trajectories of the unobserved driving styles have the expected characteristics. In the following, we will elaborate on our experiment procedures and the corresponding results. 



\subsection{Data Preparation}
To calibrate the IDM with time-varying parameters and train the NP-based model, we use the dataset \textit{pNEUMA}~\cite{pneuma}, which records the vehicle trajectories in an urban area of Athens, using ten drones from Monday to Friday with 2.5 hours per day. It includes diverse urban traffic scenarios, including normal urban lanes, interactions, and bus stops, with multiples types of vehicles.
We extract CF trajectories of cars from the raw dataset for further preprocessing. To reduce the disruptions of the lane-changing vehicles, we eliminate the trajectories $2$ s before and $2$ s after changing lanes. Moreover, the trajectories of the vehicles that stop behind intersections are also eliminated since the spacing is hard to measure. In the remaining data, we select a subset containing the data of one day with $277$ cars, where the first $200$ drivers' data are used for calibrating or training, and the rest $77$ drivers' data for testing.

\begin{figure}[t]
    \centering
    \includegraphics[width=\linewidth]{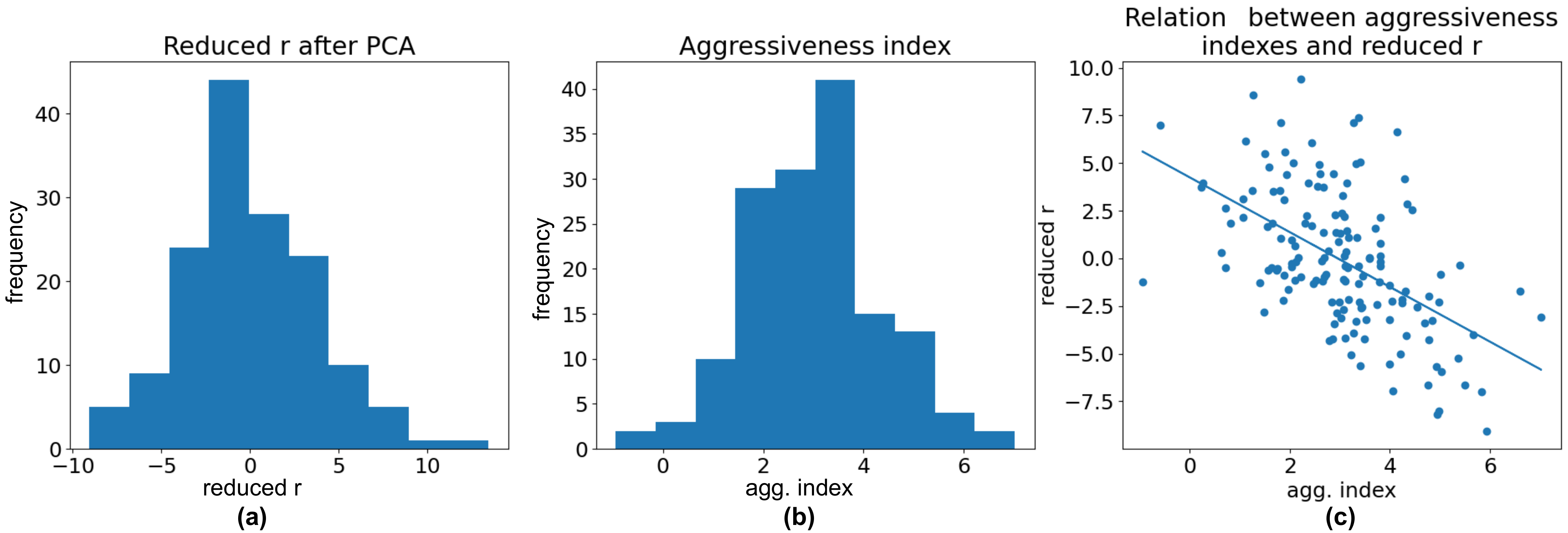}
    \caption{(a) Histogram of $\tilde{r}$; (b) histogram of $H$; and (c) relationship between $\tilde{r}$ and aggressiveness index $H$.}
    \label{fig:map_agg_2_r}
\end{figure}

\subsection{Mapping function Validation}
We use the aforementioned dataset to calculate the aggressiveness index $H$ and the reduced driving style vector $\tilde{r}$ for each driver. 
As shown in the histograms in Figure~\ref{fig:map_agg_2_r}, both $H$ and $\tilde{r}$ follow Gaussian-like distributions. Moreover, $\tilde{r}$ has a clear linear relation with $H$, and their Pearson correlation coefficient is $0.55$. Thus, $f_{map}$ is fitted using the form of a linear function. The reconstruction RMSE between $r$ and reconstructed $\hat{r}$ is around $2.51$, which is small enough to validate the feasibility of the designed mapping function~(\ref{eq:map_agg_2_r}).

\subsection{Experiment Setting}
All of the experiments are implemented using Python 3.7, PyTorch 1.9.0, and cu111 library on a machine installed with Ubuntu 18.04 and equipped with Intel Xeon 6230 CPU \nop{@ 2.10GHz}and Tesla V100S-32GB GPU.

\textbf{Calibration Algorithm Setting.} Following the guideline in~\cite{PUNZO2021103165}, we select spacing as $MoP$ and RMSE as $GoF$. The number of sample points $N$ is $5000$, the maximum iterations $M$ is $500$, and the acceptance thresholds $(\varepsilon, n_{min}, p\%)$ are $(0.01, 100, 95\%)$.

\textbf{NP Architecture and Training Setting.}
As the encoder in NP contains a deterministic path and a latent path, we define a deterministic encoder and a latent encoder correspondingly. The deterministic encoder contains $3$ hidden layers, where each layer has $128$ neurons, and the output layer has $5$ neurons as $r\in \mathbb{R}^5$. The latent encoder contains $2$ hidden layers and $2$ separate output layers to output $\mu$ and $\sigma$ of latent variable $z$. Each of the $2$ hidden layers has $5$ neurons, and each of the $2$ output layers has $1$ neuron to output the $1$-dimension $\mu$ and $\sigma$, separately. The decoder contains $3$ hidden layers, where each layer has $128$ neurons. To output $\mu$ and $\sigma$ of the predicted value, the decoder contains two separate output layers, and each has $1$ neuron since the acceleration is $1$-dimension. We utilize a customized activation function for the decoder to clip the value of output $\mu(\hat{y}_T)$, i.e., acceleration, to range $[LB_a, UB_a]$, which is formulated as $\mu(\hat{y}_T)=\mathrm{clip} \left( \mu(\hat{y}_T), LB_a, UB_a \right)$~\cite{wgan_pmlr-v70-arjovsky17a}. For training, we use Adam~\cite{kingma2014adam} as the optimizer, and the learning rate is initialized as $0.001$. The learning rate decays $90\%$ every $50$ epochs, and the training epoch is set to $200$.


\subsection{Numerical Results}
In the rest of this section, we first compare the calibration and the prediction performance of our proposed approaches with several baselines by separately computing the spacing RMSE of each driver. \nop{We compute the spacing RMSE of each driver for comparisons.}Then, we show the simulated trajectories and compare them with the ground truth. 

\textbf{Calibration Performance.}
To assess the calibration performance of our proposed IDM with time-varying parameters, we choose two models as baselines. One is the original IDM calibrated using the IP algorithm~\cite{IP_SHARMA201949} and the other one is the improved IDM LSTD-IDM~\cite{ltst2020102698}. LSTD-IDM uses the changes in some of the IDM parameters to characterize the fluctuation of human CF strategies.
Given the calibrated results, we further demonstrate the correlations between each pair of parameters to validate our design of aggressiveness index computation.

\begin{figure}[t]
    \centering
    \includegraphics[width=\linewidth]{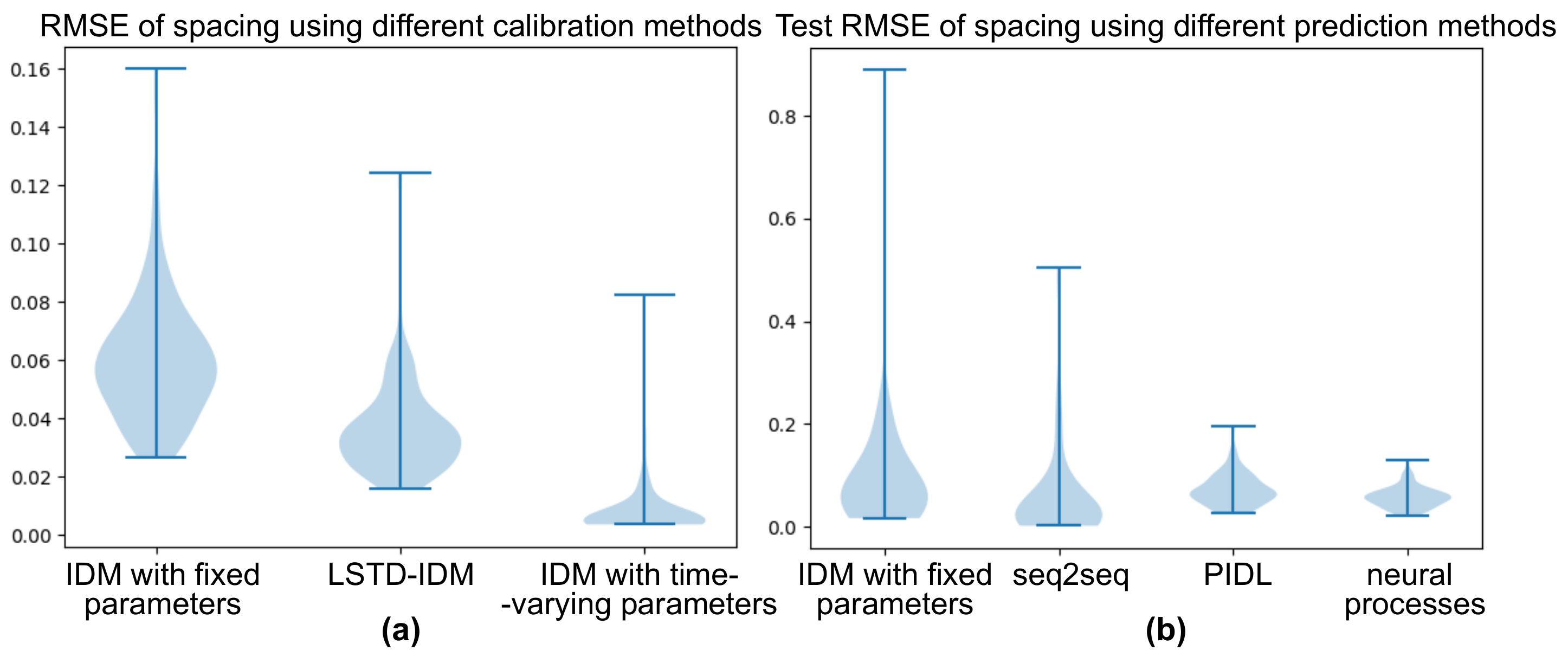}
    \caption{Comparison of the calibration and prediction RMSE of spacing using different methods.}
    \label{fig:comp_rmse}
\end{figure}

We use the mean of $\pi(\theta_t)$ as the parameters at time $t$ to calculate the RMSE of our proposed IDM with time-varying parameters. As shown in the violin plots in Figure~\ref{fig:comp_rmse}.(a), our calibration algorithm decreases the overall RMSE of $200$ drivers compared to the other two baseline models. Specifically, the mean values of the RMSE for IDM with time-varying parameters, LSTD-IDM, and the original IDM are $0.0096$, $0.0366$, and $0.0601$, respectively. Such results indicate that our calibrated time-varying parameters can better characterize the driver heterogeneity. 

\begin{figure}[t]
    \centering
    \includegraphics[width=\linewidth]{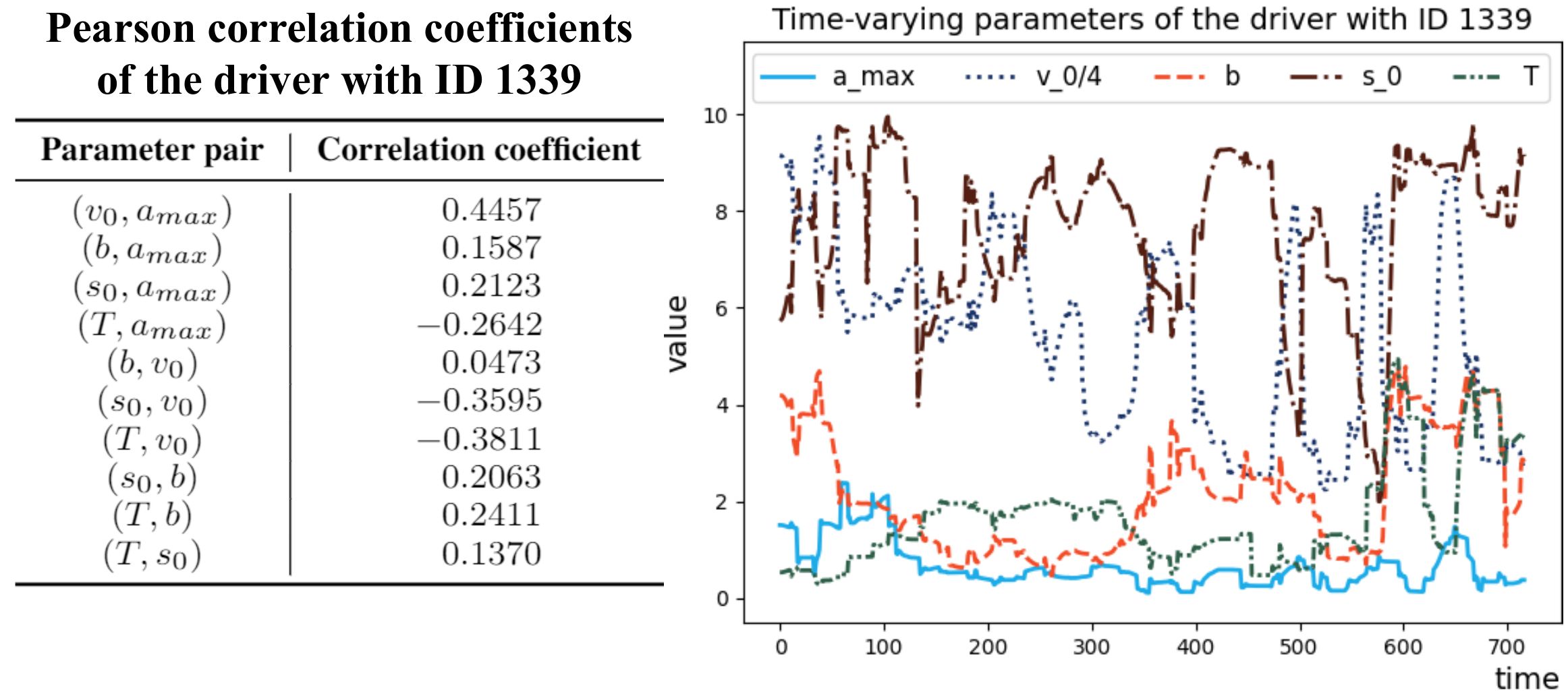}
    \caption{An example of calibrated time-varying IDM parameters and their Pearson correlation coefficients.}
    \label{fig:tv_params}
\end{figure}

Given the calibrated time-varying parameters of these $200$ drivers, Figure~\ref{fig:tv_params} illustrates the time-varying parameters of a randomly selected driver. We compute the Pearson correlation coefficient of each parameter pair for this driver. As shown in the table in Figure~\ref{fig:tv_params}, $v_0$ is positively associated with $a_{max}$ and negatively associated with $s_0$ and $T$. Namely, a high value of $v_0$, a high value of $a_{max}$, a small value of $T$, or a small value of $s_0$ can all represent a high aggressiveness level. The correlation between some CF model parameters revealed in our analysis is also consistent with the CF modeling literature~\cite{coeff_corr,IP_SHARMA201949}. Such results confirm that our assumptions in Table~\ref{tab:agg_idx} are reasonable.

\textbf{Prediction performance.}
As for the prediction performance, we compare our NP-based CF model with the original IDM, a data-driven CF model seq2seq~\cite{lstm2020102785}, and a hybrid CF model PIDL~\cite{pid2021103240}. \nop{Specifically, the former utilizes the calibrated fixed parameters to predict the future CF behaviors, while the latter utilizes recurrent neural network to learn the mapping function between the input traffic conditions and the output CF strategy.} 
Since our NP-based CF model outputs a distribution and its loss function contains a log probability, it is impractical to use the RMSE of spacing to optimize the neural network directly. Instead, we use the acceleration computed by the calibrated IDM with time-varying parameters to train our model, the seq2seq model, and the PIDL model. The detailed network architectures of seq2seq and PIDL are implemented following the experiment settings in~\cite{lstm2020102785} and~\cite{pid2021103240}, respectively. The ratio of the collocation points for training the neural network in the PIDL model is $25\%$. The evaluation procedure is conducted as follows. For an individual driver in the test set, the first $80\%$ data is used as the context points in the NP-based CF model and the points for calibrating in the other models. The rest $20\%$ data is used for comparing the performances. 

The comparison results are shown in Figure~\ref{fig:comp_rmse}.(b), where the mean of the test RMSE using our NP-based CF model is $0.0581$, the one using the original IDM is $0.0991$, the one using the seq2seq model is $0.0592$, and the one using the PIDL model is $0.0726$. Although in terms of the mean of the RMSE, our NP-based CF model only slightly outperforms the seq2seq model, the predictions of our NP-based CF model have a much smaller variance, indicating that our proposed approach can provide more stable predictions. 
Our model and the seq2seq model perform better than the other two baselines through integrating the driver heterogeneity.

\textbf{Trajectory Simulation and Road Safety Analysis.}
Besides the above comparisons, we further simulate trajectories for both observed and unobserved driving styles. The simulated trajectories of the following vehicle are compared with the ground truth in terms of the distributions of spacing, speed, and time-to-collision (TTC). Specifically, spacing and speed measure the quality of the trajectory, and TTC measures the impact on road safety and can be computed using $\mathrm{TTC}_n(t) = \frac{s_n(t)}{\Delta v_n(t)}$. A negative TTC value indicates that the collision is impossible since $\Delta v_n(t) < 0$, i.e., $v_n(t) < v_{n-1}(t)$, while a positive one represents that the collision will happen after TTC seconds if the leading vehicle and the following vehicle do not change their speeds.

\begin{figure}[t]
    \centering
    \includegraphics[width=\linewidth]{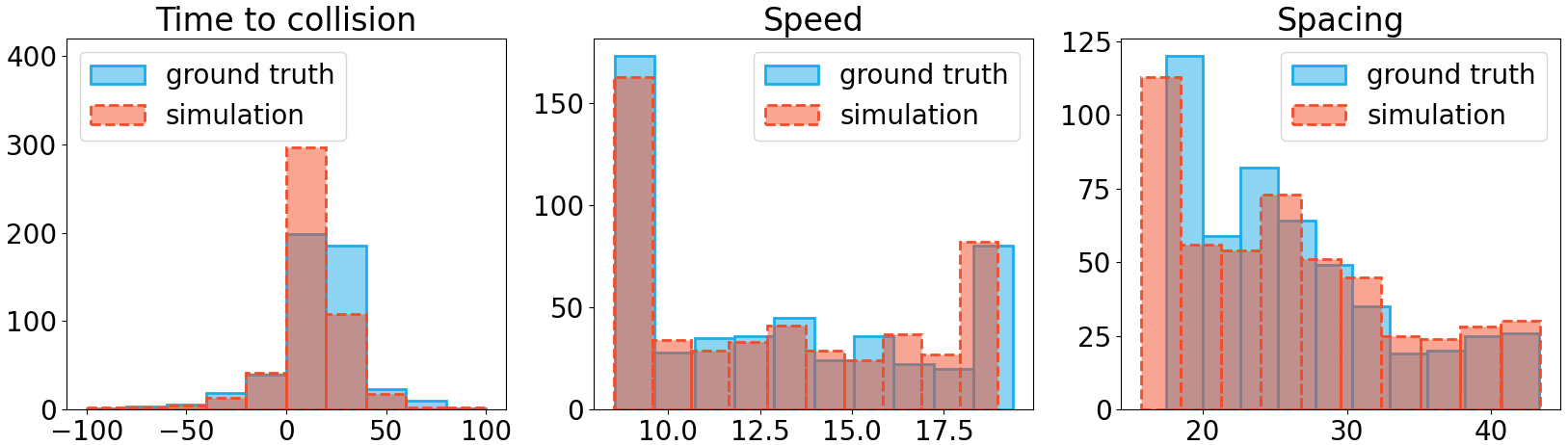}
    \caption{Histograms of TTC, speed, and spacing of the simulated trajectory and the trajectory of the ground truth.}
    \label{fig:comp_distribution}
\end{figure}

For each selected CF case, we let the leading vehicle travel the same as the ground truth, and make the following vehicle travel with the acceleration generated by our NP-based CF model. The driving style vector $r$ is derived \nop{from encoding and aggregating} using all data points of the driver. The position of the following vehicle is updated following Equation~(\ref{eq:veh_update}). The update time interval is set as the frame rate of the dataset, $0.04$ s. Due to the space limit, we only select one case for illustration. The comparison results of the driver with ID $1339$ are illustrated in Figure~\ref{fig:comp_distribution}. The similar distributions demonstrate that our NP-based CF model is capable of imitating human CF behaviors accurately in terms of the vehicle's physical movement and its resulting impact on road safety. \nop{, both in terms of the vehicle's physical movement, and its resulting impact on road safety, two primary concerns in traffic engineering in general and traffic flow modeling in particular}

\begin{figure}[t]
    \centering
    \includegraphics[width=\linewidth]{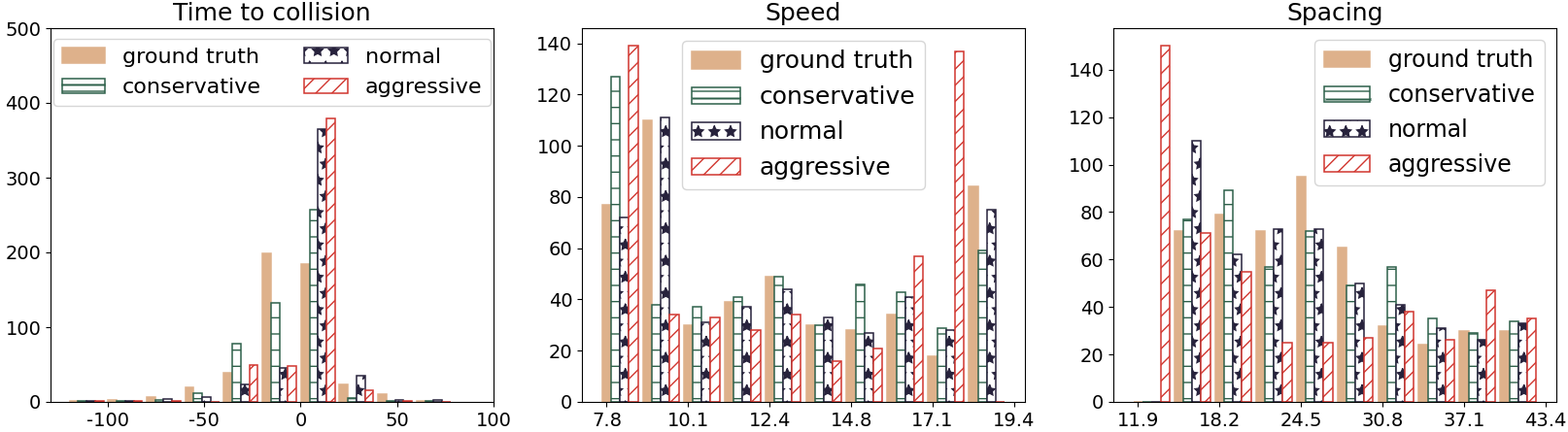}
    \caption{Histograms of TTC, speed, and spacing of the trajectory of the ground truth and the simulated trajectories of an existed driving style and two unseen driving styles.}
    \label{fig:comp_new_ds}
\end{figure}

Last but not least, we define several unobserved driving styles to simulate their corresponding CF trajectories. With the same leading vehicle, the simulated trajectory is compared with the ground truth to illustrate that the simulated trajectory fulfills the predefined driving style. During simulation, to make sure that the following vehicle with an unobserved driving style will not collide with the leading one, we define a safe distance $s_{s}$ and a brake deceleration $b_{s}$\nop{ as there is no safety constraint in the NP-based CF model}. Specifically, once the spacing is smaller than the safe distance, the following vehicle will decelerate with $b_{s}$. According to the general regulation, the safe distance at time $t$ is set dynamically as $1.5 \cdot v_n(t)$, where $1.5$ s is a typical human response time~\cite{response_time}. The typical brake deceleration of the car is $5\ \mathrm{m/s^2}$. We continue with the example of the driver with ID $1339$, whose aggressiveness index is $2.44$. Two unseen aggressiveness indexes, $0$ and $5$, are transformed to two driving style vectors. These vectors are then used as input for the decoder of the NP\nop{-based CF model} to predict accelerations of a more aggressive driving style associated with index $5$ and a more conservative driving style associated with index $0$. As there is no context point for the latent encoder to output the distribution of $z$, a normal distribution is used instead. We compare their distributions of TTC, speed, and spacing, as shown in Figure~\ref{fig:comp_new_ds}. Compared with the ground truth and the simulated trajectory of the original driving style, the generated trajectory of a more aggressive driving style (i.e., a higher aggressiveness index) has a smaller TTC, a higher speed, and a smaller spacing, while the one of a more conservative driving style has a larger TTC, a lower speed, and a larger spacing. These observations perfectly match with our expectations, demonstrating that our proposed model can generate CF behaviors for unseen driving styles.

\nop{are proposed to generate the corresponding driving style vectors $r$ and used as input for the decoder of the NP-based CF model to predict accelerations. Specifically, $0$ indicates a more conservative driving style, while $5$ indicates a more aggressive one.}

\section{Conclusion}
In response to the importance and urgency of developing traffic flow models with the capability of accounting for human factors, we have proposed a new hybrid CF model to imitate human CF behaviors of any given driving style. The proposed model combines an IDM with time-varying parameters and an NP-based CF model.
In particular, IDM with time-varying parameters is designed and calibrated to capture the driver heterogeneity when learning human CF strategies. Meanwhile, the NP-based CF model with its generative ability is applied to generate realistic human CF behaviors for any observed driving style. Furthermore, the relationship between the time-varying IDM parameters and the intermediate variable of the NP is modeled to enable the NP-based CF model to infer CF behaviors of unobserved driving styles. The experiment results demonstrate that our proposed CF model can successfully imitate CF behaviors of observed driving styles and generate CF behaviors of unobserved driving styles. Therefore, our model can be used to more realistically simulate traffic flow, investigate traffic safety, and ultimately lead to more effective traffic operation and control strategies. Our proposed model can also aid in simulating mixed traffic with automated vehicles, or in designing human-like controllers for automated vehicles.


\bibliography{aaai22}

\begin{thebibliography}{30}
\providecommand{\natexlab}[1]{#1}

\bibitem[{Arjovsky, Chintala, and Bottou(2017)}]{wgan_pmlr-v70-arjovsky17a}
Arjovsky, M.; Chintala, S.; and Bottou, L. 2017.
\newblock {W}asserstein Generative Adversarial Networks.
\newblock In Precup, D.; and Teh, Y.~W., eds., \emph{Proceedings of the 34th
  International Conference on Machine Learning}, volume~70 of \emph{Proceedings
  of Machine Learning Research}, 214--223. PMLR.

\bibitem[{Bando et~al.(1998)Bando, Hasebe, Nakanishi, and Nakayama}]{OVM}
Bando, M.; Hasebe, K.; Nakanishi, K.; and Nakayama, A. 1998.
\newblock Analysis of optimal velocity model with explicit delay.
\newblock \emph{Phys. Rev. E}, 58: 5429--5435.

\bibitem[{Barmpounakis and Geroliminis(2020)}]{pneuma}
Barmpounakis, E.; and Geroliminis, N. 2020.
\newblock On the new era of urban traffic monitoring with massive drone data:
  The pNEUMA large-scale field experiment.
\newblock \emph{Transportation Research Part C: Emerging Technologies}, 111:
  50--71.

\bibitem[{Chen et~al.(2020)Chen, Sun, Ma, Sun, and Zheng}]{ltst2020102698}
Chen, X.; Sun, J.; Ma, Z.; Sun, J.; and Zheng, Z. 2020.
\newblock Investigating the long- and short-term driving characteristics and
  incorporating them into car-following models.
\newblock \emph{Transportation Research Part C: Emerging Technologies}, 117:
  102698.

\bibitem[{Delong et~al.(2012)Delong, Osokin, Isack, and
  Boykov}]{UFLdelong2012fast}
Delong, A.; Osokin, A.; Isack, H.~N.; and Boykov, Y. 2012.
\newblock Fast approximate energy minimization with label costs.
\newblock \emph{International journal of computer vision}, 96(1): 1--27.

\bibitem[{Garnelo et~al.(2018)Garnelo, Schwarz, Rosenbaum, Viola, Rezende,
  Eslami, and Teh}]{garnelo2018neural}
Garnelo, M.; Schwarz, J.; Rosenbaum, D.; Viola, F.; Rezende, D.~J.; Eslami, S.;
  and Teh, Y.~W. 2018.
\newblock Neural processes.
\newblock \emph{arXiv preprint arXiv:1807.01622}.

\bibitem[{Gipps(1981)}]{GIPPS1981105}
Gipps, P. 1981.
\newblock A behavioural car-following model for computer simulation.
\newblock \emph{Transportation Research Part B: Methodological}, 15(2):
  105--111.

\bibitem[{Huang, Sun, and Sun(2018)}]{lstm2018car}
Huang, X.; Sun, J.; and Sun, J. 2018.
\newblock A car-following model considering asymmetric driving behavior based
  on long short-term memory neural networks.
\newblock \emph{Transportation research part C: emerging technologies}, 95:
  346--362.

\bibitem[{Huang et~al.(2018)Huang, Jiang, Zhang, Hu, Tian, Jia, and
  Gao}]{desired_thw2018194}
Huang, Y.-X.; Jiang, R.; Zhang, H.; Hu, M.-B.; Tian, J.-F.; Jia, B.; and Gao,
  Z.-Y. 2018.
\newblock Experimental study and modeling of car-following behavior under high
  speed situation.
\newblock \emph{Transportation Research Part C: Emerging Technologies}, 97:
  194--215.

\bibitem[{Jiang et~al.(2014)Jiang, Hu, Zhang, Gao, Jia, Wu, Wang, and
  Yang}]{desired_thw0094351}
Jiang, R.; Hu, M.-B.; Zhang, H.~M.; Gao, Z.-Y.; Jia, B.; Wu, Q.-S.; Wang, B.;
  and Yang, M. 2014.
\newblock Traffic Experiment Reveals the Nature of Car-Following.
\newblock \emph{PLOS ONE}, 9(4): 1--9.

\bibitem[{Kim et~al.(2019)Kim, Mnih, Schwarz, Garnelo, Eslami, Rosenbaum,
  Vinyals, and Teh}]{anp_kim2019attentive}
Kim, H.; Mnih, A.; Schwarz, J.; Garnelo, M.; Eslami, A.; Rosenbaum, D.;
  Vinyals, O.; and Teh, Y.~W. 2019.
\newblock Attentive neural processes.
\newblock \emph{arXiv preprint arXiv:1901.05761}.

\bibitem[{Kim and Mahmassani(2011)}]{coeff_corr}
Kim, J.; and Mahmassani, H.~S. 2011.
\newblock Correlated Parameters in Driving Behavior Models: Car-Following
  Example and Implications for Traffic Microsimulation.
\newblock \emph{Transportation Research Record}, 2249(1): 62--77.

\bibitem[{Kingma and Ba(2014)}]{kingma2014adam}
Kingma, D.~P.; and Ba, J. 2014.
\newblock Adam: A method for stochastic optimization.
\newblock \emph{arXiv preprint arXiv:1412.6980}.

\bibitem[{Kurtc and Treiber(2016)}]{ip2016calibrating}
Kurtc, V.; and Treiber, M. 2016.
\newblock Calibrating the local and platoon dynamics of car-following models on
  the reconstructed NGSIM data.
\newblock In \emph{Traffic and Granular flow'15}, 515--522. Springer.

\bibitem[{Lee, Ngoduy, and Keyvan-Ekbatani(2019)}]{sp_LEE2019360}
Lee, S.; Ngoduy, D.; and Keyvan-Ekbatani, M. 2019.
\newblock Integrated deep learning and stochastic car-following model for
  traffic dynamics on multi-lane freeways.
\newblock \emph{Transportation Research Part C: Emerging Technologies}, 106:
  360--377.

\bibitem[{Ma and Qu(2020)}]{lstm2020102785}
Ma, L.; and Qu, S. 2020.
\newblock A sequence to sequence learning based car-following model for
  multi-step predictions considering reaction delay.
\newblock \emph{Transportation Research Part C: Emerging Technologies}, 120:
  102785.

\bibitem[{Mo, Shi, and Di(2021)}]{pid2021103240}
Mo, Z.; Shi, R.; and Di, X. 2021.
\newblock A physics-informed deep learning paradigm for car-following models.
\newblock \emph{Transportation Research Part C: Emerging Technologies}, 130:
  103240.

\bibitem[{Ngoduy et~al.(2019)Ngoduy, Lee, Treiber, Keyvan-Ekbatani, and
  Vu}]{sp_NGODUY2019599}
Ngoduy, D.; Lee, S.; Treiber, M.; Keyvan-Ekbatani, M.; and Vu, H. 2019.
\newblock Langevin method for a continuous stochastic car-following model and
  its stability conditions.
\newblock \emph{Transportation Research Part C: Emerging Technologies}, 105:
  599--610.

\bibitem[{Osorio and Punzo(2019)}]{calibration_OSORIO2019156}
Osorio, C.; and Punzo, V. 2019.
\newblock Efficient calibration of microscopic car-following models for
  large-scale stochastic network simulators.
\newblock \emph{Transportation Research Part B: Methodological}, 119: 156--173.

\bibitem[{Punzo, Zheng, and Montanino(2021)}]{PUNZO2021103165}
Punzo, V.; Zheng, Z.; and Montanino, M. 2021.
\newblock About calibration of car-following dynamics of automated and
  human-driven vehicles: Methodology, guidelines and codes.
\newblock \emph{Transportation Research Part C: Emerging Technologies}, 128:
  103165.

\bibitem[{Saifuzzaman and Zheng(2014)}]{HF2014379}
Saifuzzaman, M.; and Zheng, Z. 2014.
\newblock Incorporating human-factors in car-following models: A review of
  recent developments and research needs.
\newblock \emph{Transportation Research Part C: Emerging Technologies}, 48:
  379--403.

\bibitem[{Sharma, Zheng, and Bhaskar(2019)}]{IP_SHARMA201949}
Sharma, A.; Zheng, Z.; and Bhaskar, A. 2019.
\newblock Is more always better? The impact of vehicular trajectory
  completeness on car-following model calibration and validation.
\newblock \emph{Transportation Research Part B: Methodological}, 120: 49--75.

\bibitem[{Sharma et~al.(2019)Sharma, Zheng, Kim, Bhaskar, and
  Haque}]{response_time}
Sharma, A.; Zheng, Z.; Kim, J.; Bhaskar, A.; and Haque, M.~M. 2019.
\newblock Estimating and Comparing Response Times in Traditional and Connected
  Environments.
\newblock \emph{Transportation Research Record}, 2673(4): 674--684.

\bibitem[{Treiber and Helbing(2003)}]{IDMM}
Treiber, M.; and Helbing, D. 2003.
\newblock Memory effects in microscopic traffic models and wide scattering in
  flow-density data.
\newblock \emph{Physical Review E}, 68(4): 046119.

\bibitem[{Treiber, Hennecke, and Helbing(2000)}]{IDM}
Treiber, M.; Hennecke, A.; and Helbing, D. 2000.
\newblock Congested traffic states in empirical observations and microscopic
  simulations.
\newblock \emph{Phys. Rev. E}, 62: 1805--1824.

\bibitem[{Yang et~al.(2019)Yang, Zhu, Liu, Wu, and Ran}]{pid8424196}
Yang, D.; Zhu, L.; Liu, Y.; Wu, D.; and Ran, B. 2019.
\newblock A Novel Car-Following Control Model Combining Machine Learning and
  Kinematics Models for Automated Vehicles.
\newblock \emph{IEEE Transactions on Intelligent Transportation Systems},
  20(6): 1991--2000.

\bibitem[{Zheng(2014)}]{lc_ZHENG201416}
Zheng, Z. 2014.
\newblock Recent developments and research needs in modeling lane changing.
\newblock \emph{Transportation Research Part B: Methodological}, 60: 16--32.

\bibitem[{Zhou et~al.(2020)Zhou, Fu, Wang, and Zhang}]{il20185034}
Zhou, Y.; Fu, R.; Wang, C.; and Zhang, R. 2020.
\newblock Modeling Car-Following Behaviors and Driving Styles with Generative
  Adversarial Imitation Learning.
\newblock \emph{Sensors}, 20(18).

\bibitem[{Zhu et~al.(2018)Zhu, Wang, Tarko, and Fang}]{IDM_best_ZHU2018425}
Zhu, M.; Wang, X.; Tarko, A.; and Fang, S. 2018.
\newblock Modeling car-following behavior on urban expressways in Shanghai: A
  naturalistic driving study.
\newblock \emph{Transportation Research Part C: Emerging Technologies}, 93:
  425--445.

\bibitem[{Zhu, Wang, and Wang(2018)}]{rl2018348}
Zhu, M.; Wang, X.; and Wang, Y. 2018.
\newblock Human-like autonomous car-following model with deep reinforcement
  learning.
\newblock \emph{Transportation Research Part C: Emerging Technologies}, 97:
  348--368.

\end{thebibliography}

\end{document}